\documentclass[letterpaper, 10 pt, conference]{ieeeconf}  % Comment this line out if you need a4paper

\usepackage{amsmath}
\usepackage{graphicx}
\usepackage{color}
\usepackage[table]{xcolor}
\usepackage{amsmath,amsfonts}
\usepackage{algorithmic}
\usepackage{algorithm}
\usepackage{array}
\usepackage[caption=false,font=normalsize,labelfont=sf,textfont=sf]{subfig}
\usepackage{textcomp}
\usepackage{stfloats}
\usepackage{url}
\usepackage{verbatim}
\usepackage{graphicx}
\usepackage{cite}
\usepackage{bm}
\usepackage{amsfonts}
\usepackage{multirow}
\usepackage{multicol}
\usepackage{booktabs}
\usepackage{algorithmic}
\usepackage{algorithm}
\usepackage{makecell}
\usepackage[T1]{fontenc}
\usepackage{lmodern} % 2026-08-26: conda texlive 缺 cm-super 的 EC/TC Type1 字体, 改用 Latin Modern (外观与 CM 几乎一致)
\usepackage[colorlinks,linkcolor=blue]{hyperref}
\DeclareUnicodeCharacter{00A0}{~}
\IEEEoverridecommandlockouts                              % This command is only needed if 
\title{\LARGE \bf
DynaSlots: Slot-Centric 3D Dynamics Pretraining for Robot Manipulation}

\author{Meng Li$^{1,2}$, Lijiang Chen$^{1}$, Qi Zhao$^{1}$, Jun Jiang$^{2}$, Longhao Zou$^{2}$, Shuchang Lyu$^{3}$, and Guangliang Cheng$^{4}$% <-this % stops a space
\thanks{Corresponding author: Jun Jiang. E-mail: jiangj@pcl.ac.cn}%
\thanks{This work was supported by Mobile Information Networks-National Science and Technology Major Project under Grant No. 2025ZD1303100}% <-this % stops a space
\thanks{$^{1}$Meng Li, Lijiang Chen, and Qi Zhao are with the Department of Electronics and Information Engineering, Beihang University, Beijing, China. E-mail: \{limenglm, chenlijiang, zhaoqi\}@buaa.edu.cn}%
\thanks{$^{2}$Meng Li, Jun Jiang, and Longhao Zou are with Pengcheng Laboratory, Shenzhen, China. E-mail: \{lim08, jiangj, Zoulh\}@pcl.ac.cn}%
\thanks{$^{3}$Shuchang Lyu is with the School of Artificial Intelligence, University of Science and Technology Beijing, Beijing, China. E-mail: lyushuchang@ustb.edu.cn}%
\thanks{$^{4}$Guangliang Cheng is with the Department of Computer Science, University of Liverpool, Liverpool, UK. E-mail: Guangliang.Cheng@liverpool.ac.uk}%
}

\begin{document}

\maketitle
\thispagestyle{empty}
\pagestyle{empty}

%%%%%%%%%%%%%%%%%%%%%%%%%%%%%%%%%%%%%%%%%%%%%%%%%%%%%%%%%%%%%%%%%%%%%%%%%%%%%%%%
\begin{abstract}
\par % DynaSlots: Slot-Centric Dynamics-Aware 3D Visual Pretraining
Robot manipulation involves localized changes: picking up a cup moves the cup while leaving the surrounding scene largely unchanged.
Yet existing 3D visual pretraining methods, including recent dynamics-aware variants, model state transitions at the scene level, compressing the motion of everything in a scene into a single global latent action.
This conflates semantically distinct transitions in scenes with multiple moving parts, obscures dependencies between changing regions, and leaves the learned representation vulnerable to task-irrelevant scene content.
We introduce DynaSlots, an action-free, label-free pretraining framework that decomposes each point cloud frame into a small set of temporally consistent slots and models dynamics at this per-slot level rather than the scene level.
A temporal slot propagation module promotes consistent slot assignments across frames; an inverse dynamics model then extracts one latent action per slot from feature differences, and a diffusion-based forward dynamics model, paired with a self-supervised interaction head, predicts each slot's future state while learning pairwise proximity-and-motion relations.
Experiments on Meta-World, Adroit, and a RealMan RM65-B robot demonstrate that DynaSlots outperforms prior scene-level baselines. It achieves superior success rates in simulation and exhibits strong robustness to scene clutter and unseen objects in real-world manipulation.
Code is available at \href{https://zhucezhanghao12345-source.github.io/DynaSlots-Anonymous/}{https://zhucezhanghao12345-source.github.io/DynaSlots-Anonymous/}.
\end{abstract}
%%%%%%%%%%%%%%%%%%%%%%%%%%%%%%%%%%%%%%%%%%%%%%%%%%%%%%%%%%%%%%%%%%%%%%%%%%%%%%%%

\section{Introduction}%%%%%slot-centric dynamics-aware 3D pretraining
\label{sec:intro}
\par
Large-scale self-supervised pretraining has become a primary lever for scalable robot learning. In manipulation, 3D representations are attractive: point clouds expose object geometry in metric space, and policies built on them~\cite{ze2024dp3,shridhar2022peract} outperform 2D counterparts with less data. This has motivated a growing line of 3D visual pretraining methods~\cite{pang2022pointmae,yu2022pointbert,zhang2022pointm2ae,zheng2024pointdif,jiang2026slp} adapting masked autoencoding~\cite{he2022mae} and cross-modal objectives~\cite{afham2022crosspoint} to point clouds, initializing policies from representations learned on large unlabeled datasets such as RH20T~\cite{fang2023rh20t} and Open X-Embodiment~\cite{embodiment2024openx}.

%% weakness
\par However, these methods largely overlook what matters most for control: dynamics. Pretraining objectives based on static appearance capture scene structure but not how it changes, and such representations underperform on manipulation benchmarks that demand understanding state transitions. Recent dynamics-aware pretraining closes this gap~\cite{liang2026afro}: an inverse dynamics model (IDM) extracts a compact latent action from feature differences across frames, and a diffusion-based forward dynamics model (FDM) predicts the next state conditioned on it, yielding substantial gains over static pretraining.

\par
\begin{figure}[t]
  \centering
  \includegraphics[width=\linewidth]{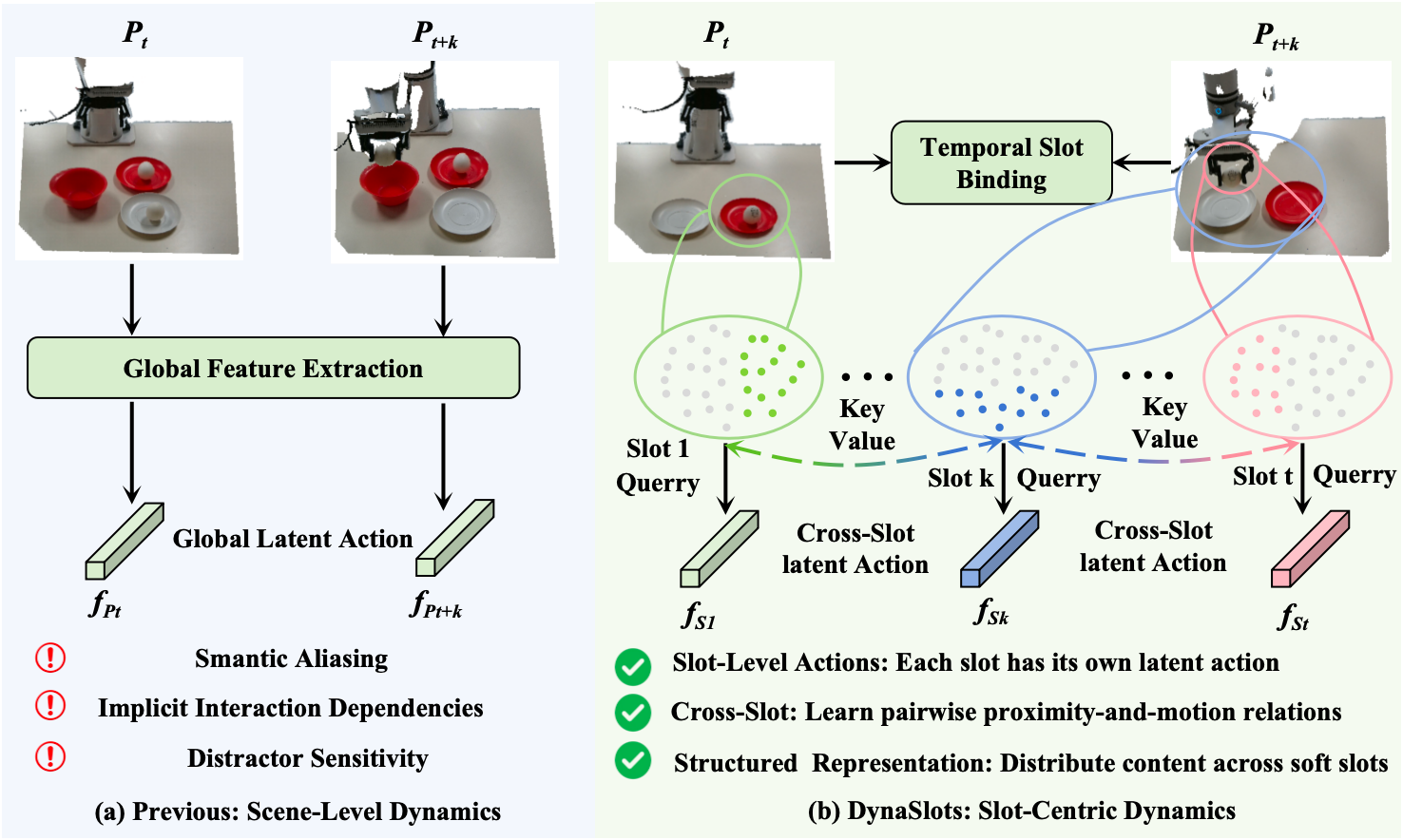}
  \caption{\textbf{Scene-level versus slot-centric 3D dynamics pretraining. }(a) Potential limitations of scene-level modeling (b) DynaSlots uses temporally aligned soft slots, assigns one latent action to each slot, and models pairwise proximity-and-motion relations across slots.}
  \label{fig:teaser}
\end{figure}

 We identify a limitation of this scene-level formulation: it summarizes an entire point cloud into one global feature and one global latent action, yet manipulation induces localized changes: when a robot picks up a milk carton, the carton moves while the table does not. This formulation presents three potential limitations (Fig.~\ref{fig:teaser}a): semantic aliasing, since similar motions in different regions can yield similar global feature differences despite representing distinct transitions; implicit interaction dependencies, since a single global vector does not explicitly represent how changes in one region relate to changes in another during manipulation; and distractor sensitivity, since task-irrelevant scene changes can perturb the global representation.

%% our approach
\par To overcome these limitations, we propose DynaSlots (Fig.~\ref{fig:teaser}b), a label-free framework for slot-level dynamics-aware 3D pretraining: the latent action of a transition should be defined per slot, not per scene. DynaSlots decomposes each frame into $K$ slots via three components: temporal slot propagation, promoting cross-frame slot consistency rather than letting independent per-frame slot attention~\cite{locatello2020slot} reorder slots arbitrarily; an inverse dynamics model extracting one latent action per slot from feature differences; and a cross-slot diffusion forward dynamics model with a self-supervised interaction head, in the spirit of neural relational inference~\cite{kipf2018nri}, that recovers a contact and co-motion relation between slots. The framework is trained end-to-end with variance-invariance-covariance regularization~\cite{bardes2022vicreg} against an EMA target encoder~\cite{caron2021dino}. In summary, the primary contributions of this paper are threefold:
\begin{itemize}
\item We introduce the first slot-based dynamics-aware 3D visual pretraining framework, requiring neither action labels, segmentation ground truth, nor reconstruction supervision.
\item We propose temporal slot propagation and a cross-slot diffusion forward dynamics model with a self-supervised interaction head to maintain cross-frame slot correspondence and capture cross-slot interactions.
\item Extensive experiments demonstrate the effectiveness of DynaSlots for robotic manipulation.
\end{itemize}

%%%%%%%%%%%%%%%%%%%%%%%%%%%%%%%%%%%%%%%%%%%%%%%%%%%%%%%%%%%%%%%%%%%%%%%%%%%%%%%%
\section{Related Work}
\label{sec:related}
\subsection{3D Visual Pretraining}
\par
Self-supervised point cloud pretraining learns transferable geometric representations from unlabeled 3D data. Inspired by masked image modeling~\cite{he2022mae}, Point-MAE~\cite{pang2022pointmae} reconstructs masked point patches, Point-BERT~\cite{yu2022pointbert} predicts discrete patch tokens, and Point-M2AE~\cite{zhang2022pointm2ae} extends masked autoencoding to multiple spatial scales. I2P-MAE~\cite{zhang2023i2pmae} further incorporates guidance from pretrained 2D models. Other objectives include diffusion-based point recovery in PointDif~\cite{zheng2024pointdif} and autoregressive patch prediction in PointGPT~\cite{chen2023pointgpt}. CrossPoint~\cite{afham2022crosspoint} aligns point clouds with rendered images through contrastive learning, while Uni3D~\cite{zhou2025uni3d} aligns 3D features with image--text representations. Structural latent points~\cite{jiang2026slp} provide a compact representation for manipulation through latent regularization and rendering supervision. These objectives emphasize spatial structure or cross-modal alignment without explicitly modeling action-dependent temporal transitions. DynaSlots instead uses per-slot dynamics as the pretraining signal for downstream manipulation.

\subsection{Dynamics-Aware Pretraining and Latent Actions}
\par
Latent-action learning extracts transition representations from videos without requiring physical action labels during pretraining. LAPO~\cite{schmidt2024lawa} learns latent actions through joint inverse and forward dynamics modeling, and LAPA~\cite{ye2025lapa} uses discrete latent actions to pretrain vision-language-action models. LAOF~\cite{bu2026laof} introduces optical-flow constraints, while Bu et al.~\cite{bu2026lam} empirically evaluate latent-action design choices. VPT~\cite{baker2022vpt} follows a different supervision regime: it trains an inverse dynamics model on action-labeled data before annotating unlabeled videos. Finer-grained alternatives include 3D motion fields extracted from human videos~\cite{yin2025motionfield}.

For representation pretraining, DynaMo~\cite{cui2024dynamo} learns from in-domain visual dynamics, and FVP~\cite{hou2025fvp} uses diffusion-based next-point-cloud prediction. AFRO~\cite{liang2026afro}, our closest baseline, jointly learns inverse and forward dynamics in a global 3D feature space without action or reconstruction supervision. Other approaches use distinct signals: DynaFLIP~\cite{lee2026dynaflip} aligns image transitions, language, and 3D flow; AEM~\cite{zhou2026aem} models vision-action histories; and Sparse2Act~\cite{guo2026sparse2act} uses end-effector actions to supervise sparse 3D features. V-JEPA~2~\cite{assran2025vjepa2} and 3D Foresight~\cite{he2026foresight} further connect visual prediction with planning or policy learning. DynaSlots builds on dynamics-aware pretraining by assigning latent actions to temporally corresponding slots and learning cross-slot relations, replacing AFRO's global transition bottleneck with a structured representation.

\subsection{Slot-Based Representation Learning}
\par
Slot Attention~\cite{locatello2020slot} maps visual features into a set of learned slots through competitive attention. SAVi~\cite{kipf2022savi} extends this mechanism to video with recurrent slot updates, and SAVi++~\cite{elsayed2022saviplus} uses depth prediction to support learning from more complex videos. SlotFormer~\cite{wu2023slotformer} learns temporal dynamics over slot sequences, while DINOSAUR~\cite{seitzer2023dinosaur} and SlotDiffusion~\cite{wu2023slotdiffusion} improve decomposition and generation through feature reconstruction and diffusion decoding, respectively. In robotic control, FOCUS~\cite{ferraro2023focus} and SOLD~\cite{mosbach2025sold} explore structured world models for model-based reinforcement learning. Neural relational inference~\cite{kipf2018nri} jointly learns latent interaction graphs and dynamics, motivating relational modeling beyond independent slot prediction.

Temporal slot consistency and relational dynamics are therefore established ideas. Our focus is their integration with per-slot latent-action learning directly on unordered 3D point clouds, using a pretraining objective that requires neither physical action labels, segmentation labels, nor reconstruction targets. Here, slots form a learned soft scene decomposition and need not correspond one-to-one with physical objects.

% ============================================================

\section{Methodology}
\label{sec:method}

\par
\begin{figure*}[t]
  \centering
  \includegraphics[width=\textwidth]{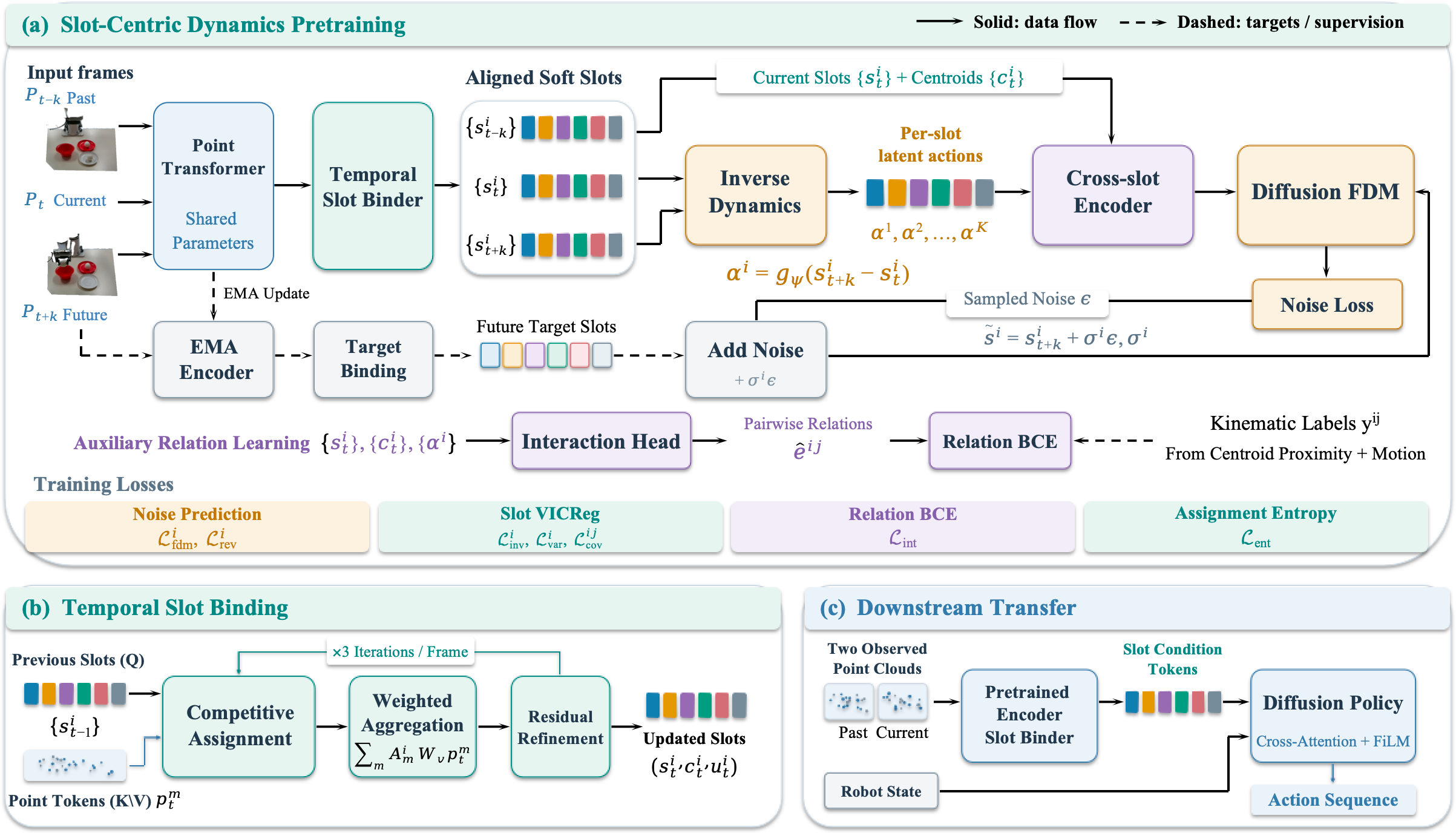}
  \caption{\textbf{DynaSlots framework.} (a) Three point-cloud frames are encoded by a shared Point Transformer and propagated by a temporal slot binder into $K{=}6$ aligned soft slots. The IDM extracts per-slot latent actions from feature differences, and the cross-slot encoder conditions a diffusion FDM on current slot features, centroids, and actions. An EMA target branch supplies noised future slot features, while an auxiliary interaction head predicts pairwise proximity-and-motion relations. The displayed losses are noise prediction, slot VICReg, relation BCE, and assignment entropy. (b) Previous slots query current point tokens through competitive assignment, weighted aggregation, and residual refinement, repeated three times per frame. (c) Two observed point clouds and the robot state produce slot condition tokens for the diffusion policy, which predicts the action sequence using cross-attention and FiLM.}
  \label{fig:method}
\end{figure*}

 Given unlabeled, coordinate-only point cloud sequences $(P_{t-k}, P_t, P_{t+k})$ sampled at stride $k$, DynaSlots learns a dynamics-aware encoder without requiring action, segmentation, or reconstruction labels. 
A Point Transformer first tokenizes each frame. A temporal binder then assigns these tokens into $K{=}6$ temporally consistent, soft slots, which capture correlated regions or parts rather than strict physical objects. 
Three modules then model per-slot dynamics: an inverse dynamics model (IDM) extracts latent actions from slot differences; a diffusion-based forward dynamics model (FDM) predicts future slot states; and an interaction head estimates pairwise proximity-and-motion relations.
The framework is optimized end-to-end using a VICReg-style loss against a momentum-updated EMA target. For downstream tasks, the pretrained encoder and binder provide slot representations that condition the diffusion policy.
Figure~\ref{fig:method} illustrates the pretraining pipeline (a), the temporal slot-binding mechanism (b), and the downstream transfer (c).

\subsection{Point Cloud Tokenization}
\par The shared tokenization path in Fig.~\ref{fig:method}(a) first maps each input point cloud to per-point features. Each point cloud is normalized per coordinate channel to $[-1,1]$ using dataset-wide limits; we do not rescale points onto a unit sphere.
DynaSlots tokenizes the full, ungrouped point set directly with a Point Transformer backbone~\cite{zhao2021point}: a linear input projection lifts each point into a $256$-dimensional working embedding, self-attention blocks let every point attend to every other point in the frame (no farthest-point sampling or neighbor grouping is used), and a final linear projection maps each per-point embedding to a token $p^m_t \in \mathbb{R}^{D}$ with $D{=}64$.
The online encoder $f_\phi$ and its exponential-moving-average target $f_\xi$ (momentum $0.996$) share this architecture and process the three frames with shared weights. The online branch provides the conditioning features, while the EMA branch supplies target features for the self-supervised losses; we write the resulting token sets generically as $T_t = \{p^m_t\}_{m=1}^{N}$ for $t \in \{t{-}k, t, t{+}k\}$.

\subsection{Temporal Slot Propagation}
\par The temporal update expanded in Fig.~\ref{fig:method}(b) is the binder used in the pretraining path of Fig.~\ref{fig:method}(a). Applying the binder independently to each frame could reorder slots over time: slot $i$ might bind to unrelated scene content at frames $t$ and $t{+}k$, causing the feature difference $s^i_{t+k} - s^i_t$ to mix assignment changes with scene dynamics.
Temporal slot propagation addresses this by using the slots of the previous frame as queries to bind the current frame, encouraging temporally consistent slot assignments. Formally, each slot $i$ at frame $t$ is a tuple $(s^i_t, c^i_t, u^i_t)$ of a content feature $s^i_t \in \mathbb{R}^{D}$, a soft centroid $c^i_t \in \mathbb{R}^{3}$, and an activation $u^i_t \in [0,1]$.
At the first frame, slots are initialized from $K$ learned queries.
For each subsequent frame, the binder scores every point against every slot and computes assignments in two stages.
Points first compete across slots,
\begin{equation}
\bar{A}^i_m = \frac{\exp\!\big(\langle W_q s^i_{t-1},\, W_k p^m_t\rangle / \sqrt{d_k}\big)}{\sum_{i'} \exp\!\big(\langle W_q s^{i'}_{t-1},\, W_k p^m_t\rangle / \sqrt{d_k}\big)},
\label{eq:binding}
\end{equation}
which sums to $1$ over slots $i$ for each point $m$; renormalizing over points, $A^i_m = \bar{A}^i_m / (\sum_{m'} \bar{A}^i_{m'} + \epsilon)$, gives the weighted-mean aggregation weight used below.
Each slot is then updated with a residual aggregation followed by a refinement step,
\begin{align}
\hat{s}^i &= s^i_{t-1} + \textstyle\sum_m A^i_m\, W_v p^m_t, \label{eq:update1}\\
s^i_t &= \hat{s}^i + \mathrm{MLP}\big(\mathrm{LN}(\hat{s}^i)\big), \label{eq:update2}\\
c^i_t &= \textstyle\sum_m A^i_m\, \mathbf{x}^m_t, \qquad u^i_t = \sigma\big(\mathrm{MLP}(s^i_t)\big). \label{eq:update3}
\end{align}
The binding step is iterated three times per frame.
During pretraining we propagate exactly two steps ($t{-}k \to t \to t{+}k$).
At inference, the binder is reinitialized from the $K$ learned queries at each policy call and propagates only over the two-frame observation window used by the downstream policy, rather than persisting across an entire rollout.

\subsection{Slot-Level Latent Actions}
\par As shown in Fig.~\ref{fig:method}(a), DynaSlots extracts one latent action per slot from temporally corresponding slot features using an inverse dynamics model:
\begin{equation}
\alpha^i = g_\psi\big(s^i_{t+k} - s^i_t\big) \in \mathbb{R}^{d}, \qquad d = 16,
\label{eq:idm}
\end{equation}
where $g_\psi$ is an MLP with LayerNorm.
Using feature differences encourages the latent action to focus on what changed and reduces direct access to static appearance in $s^i_{t+k}$.
Slots with stable content yield small feature differences, allowing the IDM to represent minimal change separately for each slot.
For reverse consistency, the same IDM maps $s^i_t - s^i_{t+k}$ to $\hat{\alpha}^i_{\mathrm{rev}}$. The FDM then predicts $s^i_t$ from $s^i_{t+k}$ and this reverse action using the noise-prediction objective in Eq.~\ref{eq:fdm}.

\subsection{Cross-Slot Diffusion Forward Dynamics}
\par The diffusion path in Fig.~\ref{fig:method}(a) predicts the future slot features $\{s^i_{t+k}\}$ from the current state $\{s^i_t\}$ and the per-slot actions $\{\alpha^i\}$. The conditioning states and actions are computed from the online branch, while the noised target features are obtained from the corresponding slots of the EMA branch.
To capture cross-slot dependencies, a three-layer Transformer applies self-attention to summaries $e^i = s^i_t + W_\alpha\alpha^i + \mathrm{PE}(c^i_t)$, where $W_\alpha \in \mathbb{R}^{D\times d}$ is a learned linear projection from $16$ to $64$ dimensions and $\mathrm{PE}:\mathbb{R}^{3}\to\mathbb{R}^{D}$ is a Fourier positional embedding. The projection and Transformer are shared across slots and prediction directions.
The generator is a diffusion transformer (DiT)~\cite{peebles2023dit} with adaptive layer-norm.
Each slot is noised independently at its own noise level $\sigma^i$, and the denoiser $\epsilon_\theta$ is trained with the simplified denoising objective~\cite{ho2020ddpm,karras2022edm}
\begin{equation}
\mathcal{L}^i_{\mathrm{fdm}} = \mathbb{E}_{\epsilon,\sigma}\Big[\, \big\| \epsilon - \epsilon_\theta\big( \underbrace{s^i_{t+k} + \sigma^i \epsilon}_{\tilde{s}^i},\, \mathrm{cond}^i,\, \sigma^i \big) \big\|^2 \Big],
\label{eq:fdm}
\end{equation}
where $\mathrm{cond}^i$ is slot $i$'s output of the condition encoder.
Each slot has its own noised target and denoising path, while its condition includes cross-slot context.
For reverse prediction, the same encoder processes $e^i_{\mathrm{rev}} = s^i_{t+k} + W_\alpha\hat{\alpha}^i_{\mathrm{rev}} + \mathrm{PE}(c^i_{t+k})$, using online future states and centroids. The shared denoiser computes $\mathcal{L}^i_{\mathrm{rev}}$ via Eq.~\ref{eq:fdm}, with the resulting reverse conditions and noised EMA-target slots at time $t$.

\subsection{Self-Supervised Interaction Head}
\par The auxiliary branch in Fig.~\ref{fig:method}(a) makes pairwise relations explicit. A lightweight head predicts, for every slot pair, the probability of a proximity-and-motion relation between their soft centroids, $\hat{e}^{ij} = \sigma(\mathrm{MLP}([s^i_t,\, s^j_t,\, c^i_t,\, c^j_t,\, \alpha^i,\, \alpha^j]))$, trained with binary cross-entropy against a purely kinematic, symmetric pseudo-label computed from the slots themselves:
\begin{align}
y^{ij} = 1 \iff\ & \big(d^{ij}_t < r_c \,\lor\, d^{ij}_{t+k} < r_c\big) \notag\\
&\;\land\; \big(\|\Delta c^i\| > v_{\min} \,\lor\, \|\Delta c^j\| > v_{\min}\big),
\label{eq:interaction-label}
\end{align}
where $d^{ij}$ is the distance between soft centroids, $r_c = 0.2\times$ the scene bounding-box diagonal, and $\Delta c^i = c^i_{t+k} - c^i_t$ is the centroid displacement with threshold $v_{\min} = 0.02$.
Intuitively, $y^{ij} = 1$ when the two slot centroids are nearby at either frame and at least one moves beyond the threshold; no annotation is required.
This symmetric label provides a slot-level proxy for contact and co-motion; it does not establish physical contact or identify the source of motion.
The interaction loss backpropagates into the shared slot and latent-action representations. The interaction head and forward dynamics model are trained jointly, and the predicted pairwise relations provide an interpretable proximity-and-motion signal.

\subsection{Training Objective}
\par The four loss groups shown at the bottom of Fig.~\ref{fig:method}(a) combine the forward and reverse noise-prediction losses with slot-level variance-invariance-covariance regularization, the interaction loss, and a slot-entropy term. We transform the raw activations into floor-normalized weights $\omega^i_t=\rho+(1-\rho)(u^i_t+\epsilon_u)/(K^{-1}\sum_j(u^j_t+\epsilon_u))$, with $\rho{=}0.5$ and $\epsilon_u{=}10^{-6}$:
\begin{align}
\mathcal{L} = {}& \lambda_{\mathrm{fdm}} \sum_i \omega^i_t \mathcal{L}^i_{\mathrm{fdm}} + \lambda_{\mathrm{reverse}} \sum_i \omega^i_t \mathcal{L}^i_{\mathrm{rev}} \notag\\
& + \sum_i \omega^i_t \big(\lambda_{\mathrm{inv}} \mathcal{L}^i_{\mathrm{inv}} + \lambda_{\mathrm{var}} \mathcal{L}^i_{\mathrm{var}}\big) + \lambda_{\mathrm{cov}} \sum_{i \neq j} \mathcal{L}^{ij}_{\mathrm{cov}} \notag\\
& + \lambda_{\mathrm{int}} \mathcal{L}_{\mathrm{int}} + \lambda_{\mathrm{ent}} \mathcal{L}_{\mathrm{ent}},
\label{eq:total}
\end{align}
with default weights $\lambda_{\mathrm{fdm}}{=}\lambda_{\mathrm{reverse}}{=}\lambda_{\mathrm{inv}}{=}\lambda_{\mathrm{var}}{=}1.0$, $\lambda_{\mathrm{cov}}{=}0.04$, and $\lambda_{\mathrm{int}}{=}\lambda_{\mathrm{ent}}{=}0.1$. Since $\omega^i_t\geq\rho$ and $K^{-1}\sum_i\omega^i_t=1$, low-activation slots receive less emphasis but cannot switch off their training signal, including the variance term.
The VICReg-style terms operate on slot contents reshaped to $[B,T,K,D]$: variance prevents collapse, covariance decorrelates slots, and invariance aligns online slot contents with the corresponding EMA-target slots.
The entropy term $\mathcal{L}_{\mathrm{ent}} = -\sum_{i,m} \bar{A}^i_m \log \bar{A}^i_m$ sharpens the per-point competition weights of Eq.~\ref{eq:binding}, encouraging each point to bind to a single slot.
We optimize with AdamW ($\mathrm{lr}{=}10^{-4}$) for $300$ epochs at frame stride $k{=}4$, pretraining separately on each task's own demonstrations rather than a shared corpus. The downstream diffusion policy is trained using slot representations from the pretrained encoder and binder, together with the robot state, as illustrated in Fig.~\ref{fig:method}(c).

% TODO(实验后): 补下游接入方式 (SlotPool / SlotSeq) 与实现细节到 Experiments; 核对超参以代码为准

%%%%%%%%%%%%%%%%%%%%%%%%%%%%%%%%%%%%%%%%%%%%%%%%%%%%%%%%%%%%%%%%%%%%%%%%%%%%%%%%
\section{Experiments}
\label{sec:ex}
\par
For each task, we train DynaSlots policies with four independent random seeds. Each policy is trained for $100$ epochs and evaluated every $10$ epochs using $25$ rollouts per checkpoint. For each seed, we select the checkpoint with the highest evaluation success rate and then average the selected scores across the four seeds. MetaWorld~\cite{yu2020metaworld} results are averaged over tasks within each difficulty group, and the overall mean is computed across all $14$ tasks. Adroit~\cite{rajeswaran2018adroit} results are averaged over the Door and Pen tasks. All results are rounded to one decimal place.

\subsection{Evaluation on Simulation}
\par
Table~\ref{tab:main} compares DynaSlots with large-scale, static-3D, policy-only, and dynamics-aware baselines~\cite{radford2021clip,oquab2023dinov2,qi2017pointnet,pang2022pointmae,zheng2024pointdif,cui2024dynamo,hou2025fvp,ze2024dp3}. All baseline results in this table are taken from AFRO~\cite{liang2026afro}.
DynaSlots achieves the highest mean success rate on both benchmarks: $80.8\%$ on MetaWorld and $88.0\%$ on Adroit, improving over AFRO by $4.8$ and $5.0$ percentage points (pp), respectively.
On MetaWorld, it reaches $92.3\%$, $74.7\%$, $60.0\%$, and $95.3\%$ in the Easy, Medium, Hard, and Very Hard groups. The corresponding gains over AFRO are $4.3$, $5.0$, $5.0$, and $4.6$ pp, showing a consistent advantage across the four difficulty groups.
On Adroit, DynaSlots achieves $88.0\%$ on both Door and Pen, exceeding AFRO by $6.0$ and $4.0$ pp, respectively.
DynaSlots also exceeds DP3 by $11.1$ pp on MetaWorld and $13.0$ pp on Adroit, and the strongest static-3D baseline on each benchmark by $16.9$ and $11.0$ pp.
These results show higher reported success rates for DynaSlots than for the compared scene-level systems.

\begin{table*}[t]
\caption{Comparison on simulation benchmarks (success rates in \%), using the evaluation schedule and baseline suite of~\cite{liang2026afro}. Parentheses give gaps to the column best in pp ($\downarrow$).}
\label{tab:main}
\centering\footnotesize
\setlength{\tabcolsep}{4.5pt}
\renewcommand{\arraystretch}{1.2}
\begin{tabular}{llcccccccc}
\toprule
\multirow{2}{*}{Category} & \multirow{2}{*}{Method} & \multicolumn{5}{c}{MetaWorld} & \multicolumn{3}{c}{Adroit} \\
\cmidrule(lr){3-7} \cmidrule(lr){8-10}
 & & Easy(3) & Med.(6) & Hard(2) & V.Hard(3) & Mean S.R. & Door & Pen & Mean S.R. \\
\midrule
\multirow{3}{*}{Large-scale} & CLIP~\cite{radford2021clip} & 29.3 & 14.0 & 5.0 & 55.3 & 24.9 ($\downarrow$55.9) & 61.0 & 84.0 & 72.5 ($\downarrow$15.5) \\
 & DINOv2~\cite{oquab2023dinov2} & 34.0 & 14.0 & 5.0 & 55.3 & 25.9 ($\downarrow$54.9) & 76.0 & 84.0 & 80.0 ($\downarrow$8.0) \\
 & PointNet~\cite{qi2017pointnet} & 71.3 & 49.7 & 23.0 & 55.3 & 51.7 ($\downarrow$29.1) & 80.0 & 72.0 & 76.0 ($\downarrow$12.0) \\
\multirow{2}{*}{Static 3D} & PointMAE~\cite{pang2022pointmae} & 77.3 & 61.3 & 42.0 & 70.0 & 63.9 ($\downarrow$16.9) & 64.0 & 76.0 & 70.0 ($\downarrow$18.0) \\
 & PointDif~\cite{zheng2024pointdif} & 83.3 & 58.0 & 41.0 & 58.7 & 61.1 ($\downarrow$19.7) & 76.0 & 78.0 & 77.0 ($\downarrow$11.0) \\
Policy-only & DP3~\cite{ze2024dp3} & 82.7 & 65.0 & 49.0 & 80.0 & 69.7 ($\downarrow$11.1) & 70.0 & 80.0 & 75.0 ($\downarrow$13.0) \\
Dyn., 2D & DynaMo~\cite{cui2024dynamo} & 42.0 & 21.7 & 14.0 & 34.0 & 27.6 ($\downarrow$53.2) & 76.0 & 68.0 & 72.0 ($\downarrow$16.0) \\
\cmidrule(lr){1-10}
\multirow{3}{*}{Dyn., 3D (scene)} & DynaMo-3D~\cite{cui2024dynamo} & 84.7 & 55.3 & 41.0 & 80.0 & 64.9 ($\downarrow$15.9) & 73.0 & 76.0 & 74.5 ($\downarrow$13.5) \\
 & FVP~\cite{hou2025fvp} & 80.0 & 44.3 & 34.0 & 62.0 & 54.3 ($\downarrow$26.5) & 66.0 & 76.0 & 71.0 ($\downarrow$17.0) \\
 & AFRO~\cite{liang2026afro} & 88.0 & 69.7 & 55.0 & 90.7 & 76.0 ($\downarrow$4.8) & 82.0 & 84.0 & 83.0 ($\downarrow$5.0) \\
\cmidrule(lr){1-10}
\rowcolor{black!8}
Dyn., 3D (slot) & DynaSlots(ours) & \textbf{92.3} & \textbf{74.7} & \textbf{60.0} & \textbf{95.3} & \textbf{80.8} & \textbf{88.0} & \textbf{88.0} & \textbf{88.0} \\
\bottomrule
\end{tabular}
\end{table*}

\begin{figure}[t]
  \centering
  \includegraphics[width=\linewidth]{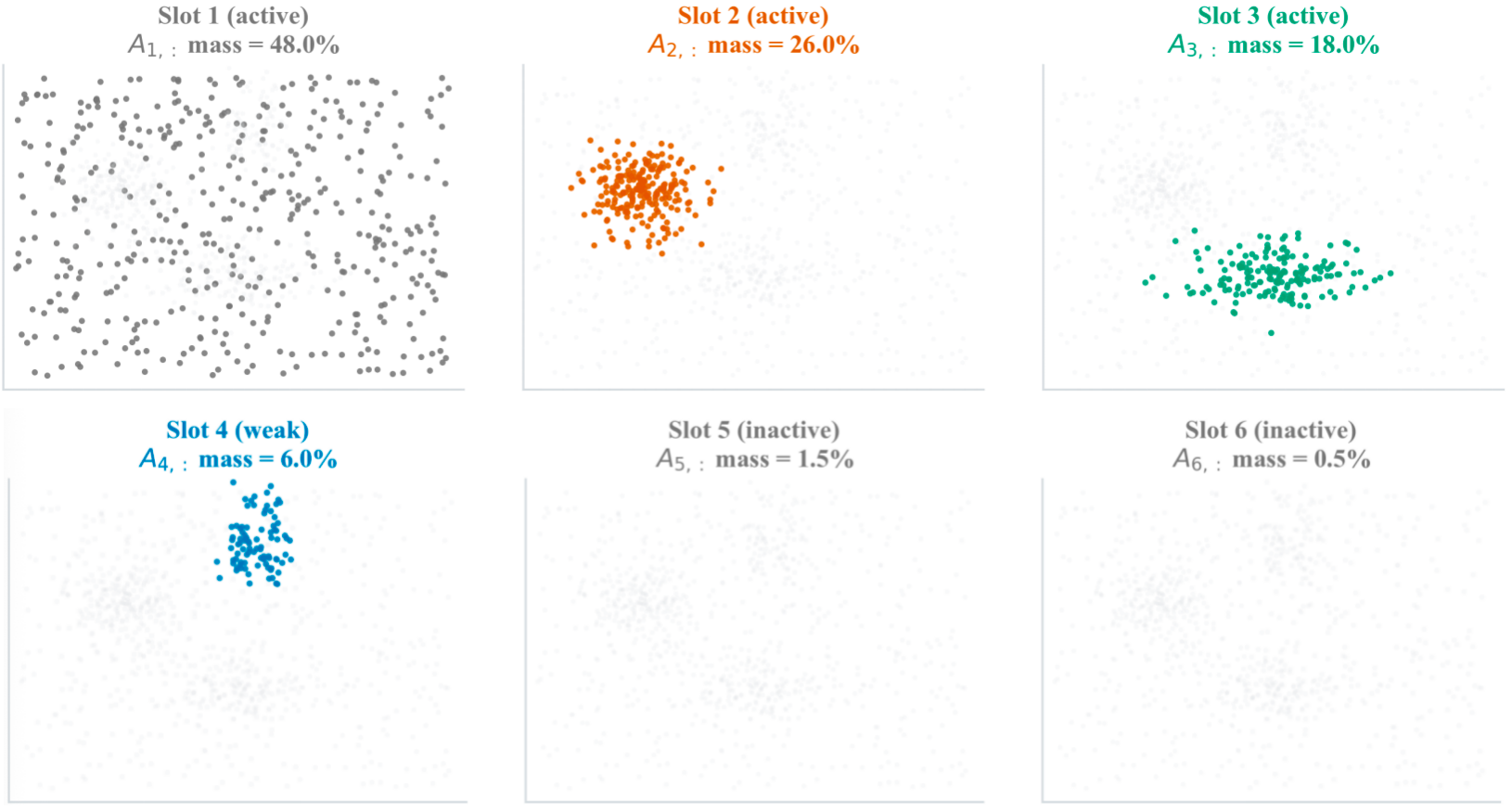}
  \caption{\textbf{Slot assignment on a representative frame ($K{=}6$).} Colored points mark where each slot has high assignment response; light-gray points are bound to other slots.}
  \label{fig:slotviz}
\end{figure}

\begin{figure*}[!t]
  \centering
  \includegraphics[width=\textwidth]{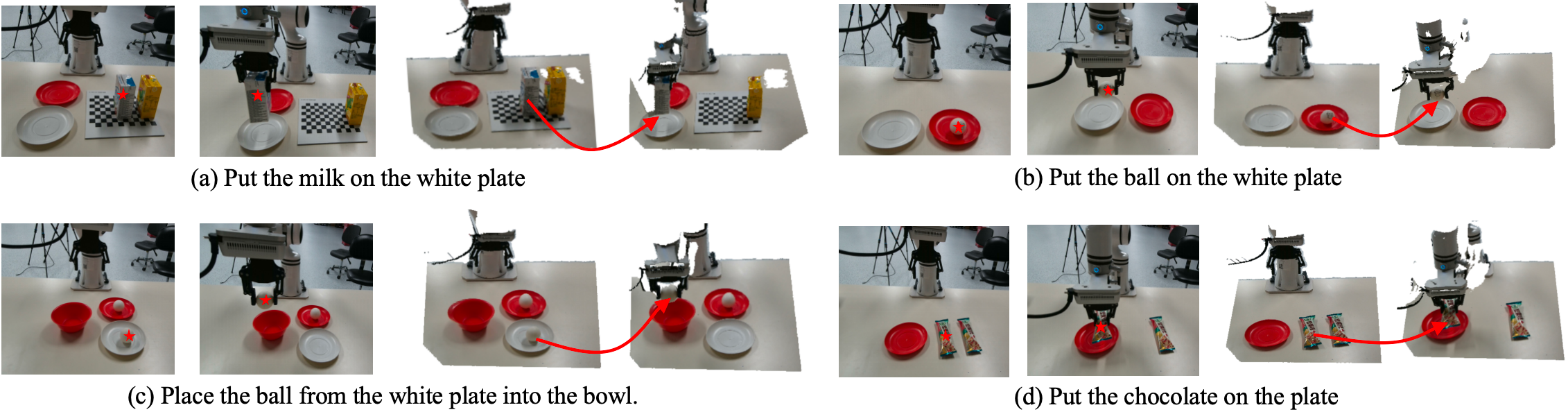}
  \caption{Four real-world manipulation tasks: (a) putting the milk on the white plate; (b) putting the ball on the white plate; (c) placing the ball from the white plate into the bowl; and (d) putting the chocolate on the plate. Each panel shows representative stages of execution.}
  \label{fig:real_tasks}
\end{figure*}

\begin{table}[t]
\caption{Ablation on DynaSlots: mean success rates (\%) and drops from the full model (pp, in parentheses).}
\label{tab:ablation}
\centering
\scriptsize
\setlength{\tabcolsep}{3pt}
\renewcommand{\arraystretch}{1.2}
\begin{tabular}{lcc}
\toprule
Variant & MetaWorld & Adroit \\
\midrule
w/o Temporal Slot Propagation
& 68.5 ($\downarrow$12.3) & 76.0 ($\downarrow$12.0) \\

w/o Interaction Head ($\lambda_{\mathrm{int}}{=}0$)
& 78.2 ($\downarrow$2.6) & 85.5 ($\downarrow$2.5) \\

w/o Reverse Consistency ($\lambda_{\mathrm{reverse}}{=}0$)
& 79.0 ($\downarrow$1.8) & 86.5 ($\downarrow$1.5) \\

$K = 4$ slots
& 74.5 ($\downarrow$6.3) & 82.5 ($\downarrow$5.5) \\

$K = 8$ slots
& 77.5 ($\downarrow$3.3) & 84.5 ($\downarrow$3.5) \\

\midrule
\rowcolor{black!8}
DynaSlots (ours, $K{=}6$)
& \textbf{80.8} & \textbf{88.0} \\
\bottomrule
\end{tabular}
\end{table}

\begin{table*}[!t]
\caption{Real-world results on tasks (a)--(d) in Fig.~\ref{fig:real_tasks}. Entries are mean success rate (\%) $\pm$ sample standard deviation (pp) across five initial-layout scores, with ten trials per layout. The Avg. is computed from the four-task suite score for each layout. The RH20T column uses RH20T pretraining followed by target-task policy training.}
\label{tab:realworld}
\centering\footnotesize
\setlength{\tabcolsep}{2pt}
\renewcommand{\arraystretch}{1.2}
\resizebox{\textwidth}{!}{%
\begin{tabular}{lccccc ccccc}
\toprule
\multirow{2}{*}{Method} & \multicolumn{5}{c}{Target-task data} & \multicolumn{5}{c}{RH20T pretrain + target-task policy training} \\
\cmidrule(lr){2-6} \cmidrule(lr){7-11}
 & Task (a) & Task (b) & Task (c) & Task (d) & Avg. & Task (a) & Task (b) & Task (c) & Task (d) & Avg. \\
\midrule
DP3~\cite{ze2024dp3} & 62.0$\pm$13.0 & 66.0$\pm$18.2 & 54.0$\pm$11.4 & 48.0$\pm$13.0 & 57.5$\pm$7.9 & 54.0$\pm$15.2 & 58.0$\pm$17.9 & 48.0$\pm$13.0 & 44.0$\pm$11.4 & 51.0$\pm$6.8 \\
PointDif~\cite{zheng2024pointdif} & 58.0$\pm$16.4 & 64.0$\pm$11.4 & 52.0$\pm$13.0 & 46.0$\pm$18.2 & 55.0$\pm$7.9 & 52.0$\pm$13.0 & 56.0$\pm$13.4 & 46.0$\pm$8.9 & 40.0$\pm$15.8 & 48.5$\pm$7.2 \\
FVP~\cite{hou2025fvp} & 60.0$\pm$10.0 & 64.0$\pm$11.4 & 54.0$\pm$8.9 & 50.0$\pm$12.2 & 57.0$\pm$6.0 & 54.0$\pm$11.4 & 60.0$\pm$12.2 & 50.0$\pm$10.0 & 42.0$\pm$13.0 & 51.5$\pm$6.0 \\
DynaMo-3D~\cite{cui2024dynamo} & 66.0$\pm$8.9 & 72.0$\pm$11.0 & 60.0$\pm$10.0 & 56.0$\pm$8.9 & 63.5$\pm$5.2 & 60.0$\pm$10.0 & 64.0$\pm$8.9 & 54.0$\pm$11.4 & 50.0$\pm$10.0 & 57.0$\pm$5.4 \\
AFRO~\cite{liang2026afro} & 72.0$\pm$11.0 & 78.0$\pm$8.4 & 68.0$\pm$8.4 & 64.0$\pm$8.9 & 70.5$\pm$4.8 & 66.0$\pm$8.9 & 72.0$\pm$11.0 & 60.0$\pm$10.0 & 54.0$\pm$8.9 & 63.0$\pm$5.4 \\
\rowcolor{black!8}
DynaSlots (ours) & \textbf{82.0$\pm$8.4} & \textbf{84.0$\pm$5.5} & \textbf{78.0$\pm$8.4} & \textbf{72.0$\pm$8.4} & \textbf{79.0$\pm$3.4} & \textbf{76.0$\pm$8.9} & \textbf{78.0$\pm$4.5} & \textbf{72.0$\pm$8.4} & \textbf{66.0$\pm$8.9} & \textbf{73.0$\pm$4.1} \\
\bottomrule
\end{tabular}}
\end{table*}

\subsection{Ablation Study}
\label{sec:ablation}
\par
Table~\ref{tab:ablation} ablates the components DynaSlots introduces (temporal slot propagation, the interaction head, reverse consistency, and the slot count $K$) on the same benchmarks as Table~\ref{tab:main}, reporting the drop from the full model (MetaWorld $80.8\%$, Adroit $88.0\%$).
Temporal slot propagation matters most by a wide margin: removing it collapses the binder to independent per-frame slot attention~\cite{locatello2020slot} and costs $12.3$ and $12.0$ pp, confirming that cross-frame slot correspondence, so that feature differences reflect temporal changes in corresponding slots, is what carries the dynamics signal.
The two auxiliary objectives each help by a smaller, consistent amount: the interaction head is worth $2.6$/$2.5$ pp and reverse consistency $1.8$/$1.5$ pp on MetaWorld/Adroit.
Finally, $K{=}6$ is the best slot count: $K{=}4$ underfits, with too few slots to separate the scene (drops of $6.3$/$5.5$ pp), while $K{=}8$ fragments it (drops of $3.3$/$3.5$ pp), so capacity should roughly match scene complexity.

\subsection{Slot Assignment Analysis}
\label{sec:slotviz}
\par
The binder's competition step (Eq.~\ref{eq:binding}) softly assigns every point token across the $K$ slots via a softmax over the slot dimension, so slots compete for points rather than each absorbing the whole scene.
Figure~\ref{fig:slotviz} visualizes this on a representative frame, coloring the points each slot claims and reporting its mean assignment mass $\bar{m}^i=\frac{1}{N}\sum_m \bar{A}^i_m$ (summing to one over slots).
Two things stand out.
The decomposition is spatial and unsupervised: active slots respond over coherent regions with no segmentation labels, purely as a byproduct of the objectives in Sec.~\ref{sec:method}.
And slot usage is adaptive: a few slots carry the scene while others stay near-inactive (small residual mass, not exactly zero), so $K{=}6$ is the binder's capacity, not a per-frame target.
The map reflects soft assignment competition, not labeled masks.

%%%%%%%%%%%%%%%%%%%%%%%%%%%%%%%%%%%%%%%%%%%%%%%%%%%%%%%%%%%%%%%%%%%%%%%%%%%%%%%%

\subsection{Real-World Experiments}
\label{sec:real}
\par
We evaluate on a RealMan RM65-B robotic arm with a parallel-jaw gripper and an Intel RealSense RGB-D camera mounted in a fixed eye-to-hand configuration. The camera extrinsics are calibrated to the robot base frame, aligning the visual observation and action spaces. Figure~\ref{fig:real_tasks} shows four real-world tasks: (a) putting the milk on the white plate, (b) putting the ball on the white plate, (c) placing the ball from the white plate into the bowl, and (d) putting the chocolate on the plate. These tasks cover pick-and-place, destination selection, and placement into a container. We collect $100$ expert trajectories for each task and use the task-specific demonstrations for pretraining and policy training.
Raw point clouds are transformed into the robot-base frame, cropped to the workspace. Each task is evaluated on five varied initial layouts, with ten trials per layout. We report the mean success rate and the sample standard deviation of the five layout-level success rates. The Avg. column first computes a four-task suite score for each layout and then averages the five suite scores.

\subsubsection{Real-World Task Performance}
\par

As shown in Table~\ref{tab:realworld}, on target-task data, DynaSlots obtains the highest success rate on all four tasks and reaches a $79.0\%$ suite average. Compared with AFRO, the gains are $10.0$, $6.0$, $10.0$, and $8.0$ pp on tasks (a)--(d), respectively, yielding an average improvement of $8.5$ pp. With RH20T pretraining followed by target-task policy training, DynaSlots remains best in every column and reaches a $73.0\%$ average, $10.0$ pp above AFRO. The shift from target-task data to RH20T pretraining reduces DynaSlots' success rate by $6.0$ pp, compared with a $7.5$ pp reduction for AFRO, indicating that the slot-centric representation preserves its advantage under a changed pretraining source.
The gains span different manipulation patterns rather than a single object or motion primitive: DynaSlots improves most on tasks (a) and (c), which require selecting a destination or transferring an object between receptacles, while it also maintains clear margins on the direct placement tasks (b) and (d). DynaSlots remains the best-performing method on all four tasks after RH20T pretraining.

\subsubsection{Cluttered-Scene Evaluation}
\par
To measure sensitivity to distractors, we evaluate task (c), placing the ball from the white plate into the bowl, with and without additional scene clutter. Table~\ref{tab:cluttergen} reuses the target-task score from Table~\ref{tab:realworld} as the matched no-clutter reference and reports five layouts with ten trials per layout in each condition. The drop is the absolute change in mean success rate, measured in pp. DynaSlots obtains $70.0\%$ in cluttered scenes, $30.0$ pp above AFRO, and its $8.0$ pp drop is the smallest among all methods; AFRO drops by $28.0$ pp. Thus, DynaSlots retains $89.7\%$ of its no-clutter success rate (70/78), whereas the strongest scene-level baseline retains $58.8\%$ (40/68).
The cluttered-scene score of DynaSlots also remains above AFRO's matched no-clutter score of $68.0\%$, whereas all other methods lose at least $24.0$ pp under clutter. This separation indicates that the slot binder can preserve task-relevant regions when additional objects change the global point-cloud content.

\begin{table}[t]
\caption{Robustness to scene clutter on task (c); drops in pp.}
\label{tab:cluttergen}
\centering\footnotesize
\includegraphics[width=0.85\linewidth]{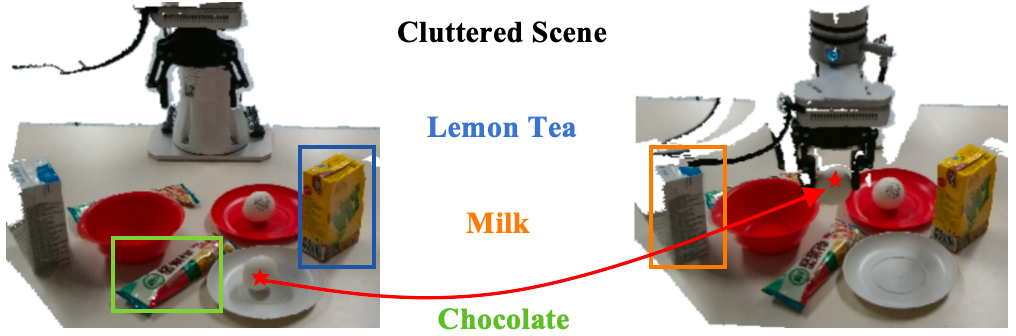}\par
\vspace{4pt}

\setlength{\tabcolsep}{2pt}
\renewcommand{\arraystretch}{1.2}
\begin{tabular}{p{\dimexpr0.34\linewidth-2\tabcolsep\relax}
>{\centering\arraybackslash}p{\dimexpr0.25\linewidth-2\tabcolsep\relax}
>{\centering\arraybackslash}p{\dimexpr0.24\linewidth-2\tabcolsep\relax}
>{\centering\arraybackslash}p{\dimexpr0.17\linewidth-2\tabcolsep\relax}}
\toprule
\multirow{2}{*}{Method} & \multicolumn{2}{c}{Success rate (\%) $\uparrow$} & \multirow{2}{*}{\makecell{Drop (pp)\\$\downarrow$}} \\
\cmidrule(lr){2-3}
 & No clutter & Cluttered & \\
\midrule
DP3~\cite{ze2024dp3} & $54.0\pm11.4$ & $28.0\pm14.8$ & $26.0$ \\
PointDif~\cite{zheng2024pointdif} & $52.0\pm13.0$ & $26.0\pm13.4$ & $26.0$ \\
FVP~\cite{hou2025fvp} & $54.0\pm8.9$ & $30.0\pm15.8$ & $24.0$ \\
DynaMo-3D~\cite{cui2024dynamo} & $60.0\pm10.0$ & $36.0\pm13.4$ & $24.0$ \\
AFRO~\cite{liang2026afro} & $68.0\pm8.4$ & $40.0\pm12.2$ & $28.0$ \\
\rowcolor{black!8}
\textbf{DynaSlots (ours)} & $\mathbf{78.0}\pm8.4$ & $\mathbf{70.0}\pm7.1$ & $\mathbf{8.0}$ \\
\bottomrule
\end{tabular}
\end{table}

\subsubsection{Unseen-Object Evaluation}
\par
We next test task (c) in scenes containing objects not encountered during training, while keeping the task goal and evaluation budget unchanged. Table~\ref{tab:objgen} compares these results with the matched original-scene scores from Table~\ref{tab:realworld}; each condition uses five layouts and ten trials per layout. DynaSlots reaches $72.0\%$ with unseen scene objects, outperforming AFRO by $18.0$ pp. Its success rate decreases by only $6.0$ pp from the original scenes, compared with $14.0$ pp for AFRO and $12.0$ pp for DynaMo-3D. This result supports improved tolerance to object-level scene variation.
This evaluation complements the clutter test by replacing scene content rather than simply adding distractors. DynaSlots retains the highest success rate after the change and has the smallest reduction among all methods, indicating that its advantage is linked to stable slot-level dynamics features instead of memorization of the training objects. Across the three real-world evaluations, DynaSlots is the best-performing method in every reported condition. Its slot-centric representation shows smaller losses under clutter and remains effective when scene objects differ from those used during training.

\begin{table}[t]
\caption{Robustness to unseen objects on task (c); drops in pp.}
\label{tab:objgen}
\centering\footnotesize
\includegraphics[width=0.85\linewidth]{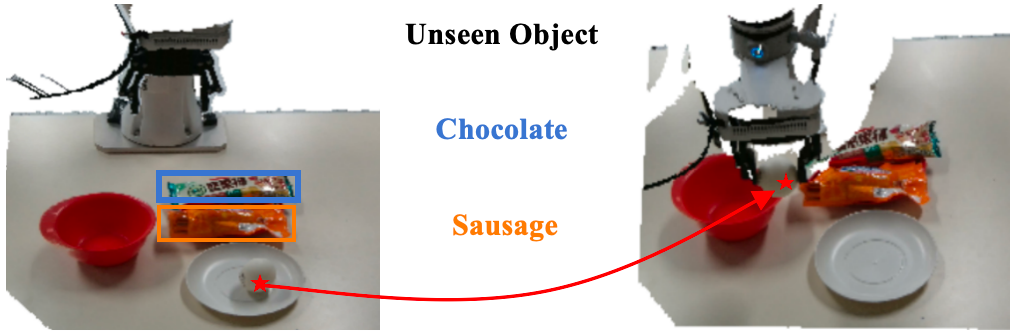}\par
\vspace{4pt}

\setlength{\tabcolsep}{2pt}
\renewcommand{\arraystretch}{1.2}
\begin{tabular}{p{\dimexpr0.34\linewidth-2\tabcolsep\relax}
>{\centering\arraybackslash}p{\dimexpr0.25\linewidth-2\tabcolsep\relax}
>{\centering\arraybackslash}p{\dimexpr0.24\linewidth-2\tabcolsep\relax}
>{\centering\arraybackslash}p{\dimexpr0.17\linewidth-2\tabcolsep\relax}}
\toprule
\multirow{2}{*}{Method} & \multicolumn{2}{c}{Success rate (\%) $\uparrow$} & \multirow{2}{*}{\makecell{Drop (pp)\\$\downarrow$}} \\
\cmidrule(lr){2-3}
 & Original & Unseen & \\
\midrule
DP3~\cite{ze2024dp3} & $54.0\pm11.4$ & $38.0\pm14.8$ & $16.0$ \\
PointDif~\cite{zheng2024pointdif} & $52.0\pm13.0$ & $36.0\pm13.4$ & $16.0$ \\
FVP~\cite{hou2025fvp} & $54.0\pm8.9$ & $40.0\pm10.0$ & $14.0$ \\
DynaMo-3D~\cite{cui2024dynamo} & $60.0\pm10.0$ & $48.0\pm11.0$ & $12.0$ \\
AFRO~\cite{liang2026afro} & $68.0\pm8.4$ & $54.0\pm11.4$ & $14.0$ \\
\rowcolor{black!8}
\textbf{DynaSlots (ours)} & $\mathbf{78.0}\pm8.4$ & $\mathbf{72.0}\pm8.4$ & $\mathbf{6.0}$ \\
\bottomrule
\end{tabular}
\end{table}

%%%%%%%%%%%%%%%%%%%%%%%%%%%%%%%%%%%%%%%%%%%%%%%%%%%%%%%%%%%%%%%%%%%%%%%%%%%%%%%%
\section{Conclusion}
\label{sec:conclusion}
\par
We introduced DynaSlots, a slot-centric, dynamics-aware 3D pretraining framework that learns temporally consistent slots, per-slot latent actions, and cross-slot interactions without action, segmentation, or reconstruction labels. Across simulation and real-world manipulation benchmarks, DynaSlots consistently outperforms scene-level baselines, demonstrating the value of slot-level dynamics modeling. Future work will evaluate more embodiments and extend compositional-generalization studies beyond the current placement setting.

\bibliographystyle{IEEEbib}

\AddToHook{env/thebibliography/begin}{\AddToHookNext{cmd/list/after}{\setlength{\itemsep}{0pt}}}
\bibliography{refs}

\end{document}